# BADREX: In situ expansion and coreference of biomedical abbreviations using dynamic regular expressions

Phil Gooch[1]

[1]Centre for Health Informatics, City University London, UK

Philip.Gooch.1@city.ac.uk

**ABSTRACT**

BADREX uses dynamically generated regular expressions to annotate term definition–term abbreviation pairs, and corefers unpaired acronyms and abbreviations back to their initial definition in the text. Against the Medstract corpus BADREX achieves precision and recall of 98% and 97%, and against a much larger corpus, 90% and 85%, respectively. BADREX yields improved performance over previous approaches, requires no training data and allows runtime customisation of its input parameters.

BADREX is freely available from https://github.com/philgooch/BADREX-Biomedical-Abbreviation-Expander as a plugin for the General Architecture for Text Engineering (GATE) framework and is licensed under the GPLv3.

## 1 INTRODUCTION

Identification of abbreviations and acronyms, or *short forms* (SF), for given term definitions, or *long forms* (LF), is a well researched topic in the biomedical natural language processing domain (see Torii *et al.* 2007 for a review). Existing tools such as Schwartz & Hearst (2003) and Ao & Takagi (2005) extract LF-SF pairs for dictionary creation, but do not provide automatic expansion of SFs within the text at the point at which they occur, which is a necessary precursor for semantic type assignment and coreference resolution. BADREX identifies, expands and annotates LF-SF pairs, and coreferences subsequent SF mentions back to their most recent definition in the text. This may facilitate disambiguation of unpaired abbreviations not possible with dictionary lookup alone (Stevenson *et al.* 2009).

## 2 METHODS

### 2.1 BADREX development

BADREX is implemented in Java as a plugin for the General Architecture for Text Engineering (GATE) framework (Cunningham *et al.* 2002). It takes a `Set` of sentences from GATE's sentence splitter, over which it iterates once. For each sentence, five processing steps are performed, where Step 1 is similar to the first stage outlined in Schwartz & Hearst (2003) and Step 2 to the third phase of Ao & Takagi (2005):

(1) Identification of candidate `<LF, SF>` pairs
(2) Applying discard conditions to `<LF, SF>` candidates to filter unwanted pairs
(3) Identifying the shortest substring in `LF` that best matches `SF` given the constraints of Steps 1 and 2
(4) Matching characters in `SF` against characters in `LF`
(5) Optional coreference of unpaired abbreviations/acronyms that match previously found short forms

In Step 1, we create two patterns: the 'head' regular expression (regex) identifies a string that contains `{1, maxOuterWords}` words followed by a string of `{1, maxInnerChars}` characters in parentheses or square brackets, and where the first character of the first group is an alphanumeric that matches the first character of the second group. The 'tail' regex consists of a similar pattern but where the first character of the *last* word of the first group is an alphanumeric that matches the *last* character of the second group. For each sentence in the input, if no match is made by the first pattern, then the second pattern may be executed. The 'head' pattern will identify candidate pairs such as:

<u>t</u>he behaviour of confluent SV40 transformed rabbit corneal epithelial cells (<u>t</u>RCEC)     (1)

and the 'tail' pattern identifies pairs such as

with two-dimensional proton nuclear magnetic <u>r</u>esonance (2D 1H NM<u>R</u>)     (2)

(matching characters underlined). In simplified form, the 'head' pattern can be expressed as:

`\b((\w)\W{0,2}(\w+\W?){1,maxOuterWords})\s*`
`\((\2.{1,maxInnerChars})(\p{Punct}\s*\w+)?\)`

and the 'tail' pattern as

`\b(.{1,maxOuterChars}\b(\w)(\w+\W?))\s*`
`\((.{1,maxInnerChars}\2(\p{Punct}\s*\w+)?)\)`

where *maxOuterWords* is the value of the user-defined parameter for the maximum number of words in the long form (default: 10, as per Ao & Takagi 2005), *maxInnerChars* is the maximum number of characters in the short form (default: 40, i.e. 10 words), and *maxOuterChars = maxOuterWords * 4*.

Usually, the short form will appear in parentheses following the long form, but they may appear in reverse order. We allow for this by setting the maximum number of characters as the same by default in both `LF` and `SF`. If the matched short form is longer than the candidate long form text preceding it, the values of `LF` and `SF` are swapped, so that `SF` always points to the abbreviation/acronym, and `LF` always to the definition.

In Step 2 we make use of a simplified subset of the discard conditions for short forms given in Appendix 1 of Ao & Takagi (2005). For example, short forms starting with a preposition, or starting and ending with a digit, are discarded. These conditions are implemented as regular expressions loaded from external configuration files, allowing this behaviour to be easily customised.

In Step 3, we use dynamically generated regular expressions to find the shortest substring of `LF` following a preposition (if present) and where *either* the first character matches the first character of `SF`, *or* the first character of the last word matches the last character of `SF`, depending on whether the 'head' or 'tail' pattern was executed in Step 1. In example (1) above, '*the behaviour of confluent SV40 transformed rabbit corneal epithelial cells*' would be shortened to '*transformed rabbit corneal epithelial cells*'.

In Step 4, we strip non-alpha characters from `LF` and `SF`, split `LF` into a character array, and iterate over `SF` to match adjacent characters, in the same order, in the `LF` array. If the proportion of matches in relation to the total alpha characters in `SF` >= *threshold* (default: 0.80), then the `<LF, SF>` pair is accepted and added to a `Map` of `<SF, LF>` key/value pairs.

The accepted pair are converted to inline annotations in the text by making use of the `start()` and `end()` methods of the regex `Matcher`, adjusted for term truncation in Step 3 and for the start offset of each sentence. The value of `LF` is stored as a feature on `SF`, and vice versa. If the term definition has already been annotated with one of a configurable set of known annotations, this annotation is used. For example, if '*transformed rabbit corneal epithelial cells*' was previously annotated as `AnatomicalSite`, then '*tRCEC*' would also be annotated with this semantic type.

In Step 5, we generate regex `Matchers` over the `Map` of pairs populated up to that point, and use these to locate and annotate unpaired, candidate SFs in sentences forward of the point at which the corresponding LF-SF pair was first introduced.

Figures 1 and 2 show sample BADREX output in the GATE Developer application for two abstracts from the evaluation corpora. Figure 1 shows semantic type assignment (*DiseaseOrSyndrome* and *Protein*) copied from the long form to subsequent short form mentions, and coreference and expansion of unpaired short forms. In coreference mode, short forms occurring within subsequent long forms are also expanded: here, the 'WAS protein' term contains an inner 'WAS' abbreviation that has been expanded to 'Wiskott-Aldrich syndrome'.

Figure 2 shows how BADREX allows for whitespace variations in subsequent mentions of the earlier-introduced short form. Future versions of BADREX may incorporate configurable string distance metrics for greater control of coreference resolution of unpaired short forms.

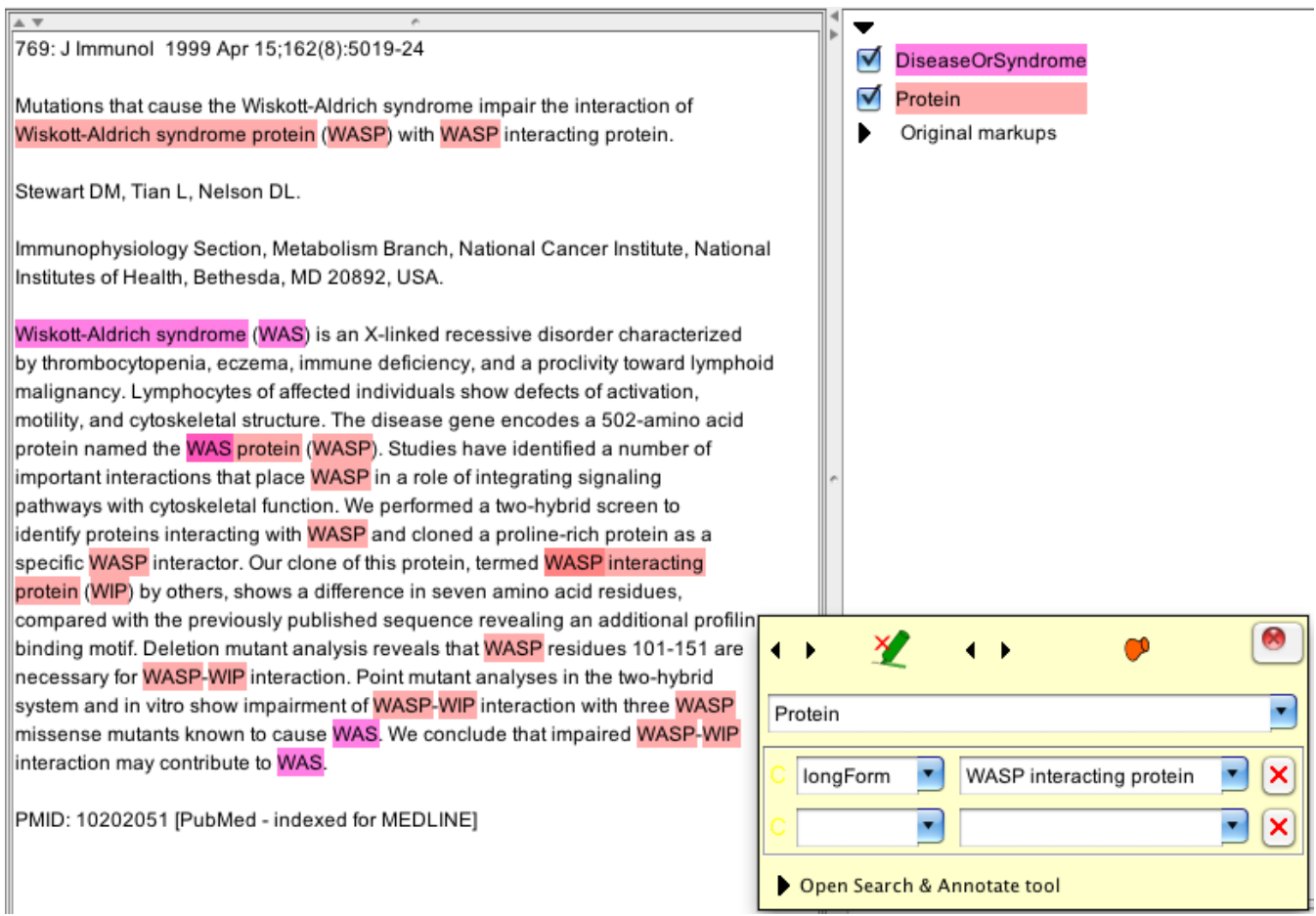

**Fig. 1.** Visualisation of BADREX output in GATE, showing automatically annotated and coreferenced short forms.

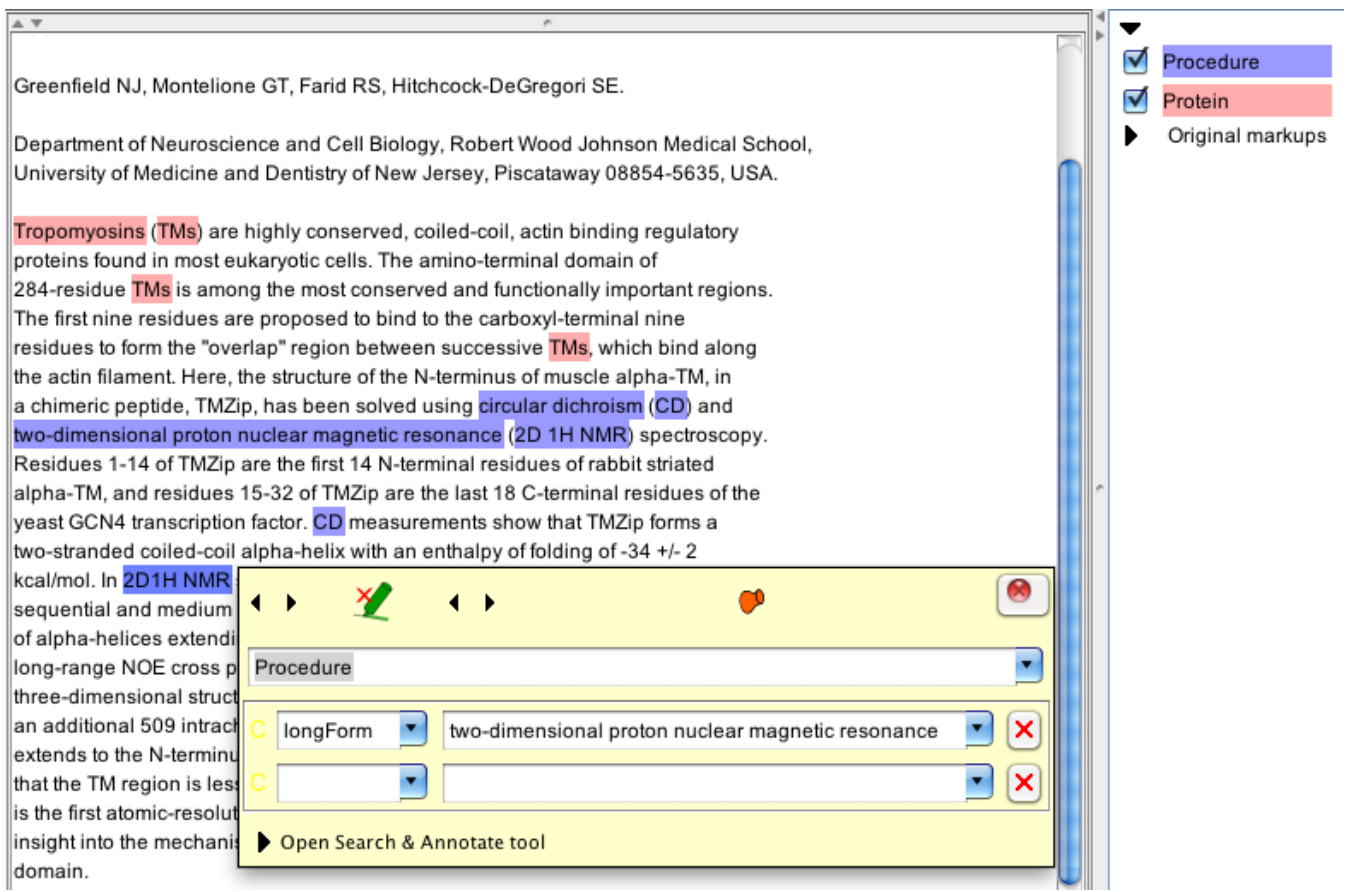

**Fig. 2S.** Example showing how BADREX's abbreviation coreference allows for white-space variations in subsequent mentions of the initially introduced short form. Here, '2D1H NMR' is coreferenced with '2D 1H NMR' and annotated with the original long form text as a feature.

### 2.2 Correction of BioText and Medstract data

The labelled, gold standard BioText 'yeast' data (http://biotext.berkeley.edu/data.html; Schwartz & Hearst 2003) comprises 1000 MedLine abstracts in a plain text file containing 954 LF-SF pairs annotated with XML-like tags, e.g.

```
<Long id=1>endoplasmic reticulum</Long>
(<Short id=1>ER</Short>)
```

where the 'id' attribute on the <Long> element matches that in the corresponding <Short> element. Using a standard XML parser, we identified and corrected errors in malformed 'id' attributes and mismatched or malformed <Long> and <Start> tags. Correction iterations continued until the file parsed.

The Medstract corpus (Pustejovsky *et al*. 2002; (http://www.medstract.org/index.php?f=gold-standard) comprises 400 MedLine abstracts in a plain text file, where 414 gold standard LF-SF pairs have been extracted into a separate text file (http://www.medstract.org/index.php?f=gold-result; the 'markables'). We analysed the markables file for offset errors, and following correction of these, we compared the abstracts file against the markables to identify any missing pairs.

We evaluated the performance of BADREX against the corrected BioText and Medstract corpora, and compared the performance alongside that of three published systems: Schwartz & Hearst (S-H, 2003), ALICE (Ao & Takagi 2005) and MBA (Xu *et al*. 2009) against the same data. For S-H and ALICE, executable code was available to evaluate on the corrected corpora; for MBA, code was not available so we report the Medstract figures provided by Xu *et al*.

**Table 1.** Missing and corrected short-form-long-form pairs in the Medstract gold standard markables

| Short form | Long form |
|---|---|
| *Missing pairs* | |
| hCG | human chorionic gonadotrophin |
| eNOS | endothelial type of NO synthase |
| 3beta-HSD II | 3beta-hydroxysteroid dehydrogenase type II |
| tTGase | tissue transglutaminase |
| hMG | human menopausal gonadotrophin |
| IVF ET | in vitro fertilization/embryo transfer |
| hMG | human menopausal gonadotrophin |
| hCG | human chorionic gonadotrophin |
| ds | double-stranded |
| frag | fragmentation |
| 3-D | 3-dimensional |
| 22K hGH | 22 kDa growth hormone |
| alpha-DB | alpha-dystrobrevin |
| bHLH | basic helix-loop-helix |
| b FGF | basic fibroblast growth factor |
| CI | confidence interval |
| oc | Osteosclerosis |
| topo II | topoisomerase II |
| ALP | alkaline phosphatase levels |
| BMD | bone mineral density |
| CI | confidence interval |
| micro-CT | micro-computed tomography |
| PrE | primitive endoderm |
| hHb1 | Human hair keratin basic 1 |
| bp | base pair |
| mtDNA | mitochondrial genome |
| beta 2M | beta 2-microglobulin |
| pb | peripheral blood |
| AT | Ataxia teleangiectasia |
| I.L.S.G. | International Lymphoma Study Group |
| R.E.A.L. | Revised European-American Classification of Lymphoid Neoplasms |
| tHcy | total homocysteine |
| iNOS | inducible nitric oxide synthase |
| 5-FU | 5-fluorouracil |
| rAAV | recombinant adeno-associated virus |
| oriP | origin of latent viral DNA replication |
| HVJ | hemagglutinating virus of Japan |
| E0' | equilibrium reduction potential |
| O2- | superoxide |
| eNOS | endothelial NO synthase |
| GlOx | glutamate oxidase |
| beta-END | beta endorphin |
| tRCEC | transformed rabbit corneal epithelial cells |

| Short form | Corrected long form | Original long form |
|---|---|---|
| *Corrected pairs* | | |
| RAR | RA receptor | regulation of tissue transglutaminase |
| IAA | indoleacetic acid | in the presence of 10(-6) m 3-indoleacetic acid |
| EXACCT | exonuclease-amplification coupled capture technique | e exonuclease-amplification coupled capture technique |
| GlyRalpha2 E3A | glycine alpha2 exon 3A | glycine alpha2 exon 3a (glyralpha2 e3a) and gaba(a) exon gamma |
| EGFr | EGF receptor | eration through binding to egf receptor |
| VIN | vulval intraepithelial neoplasia | val intraepithelial neoplasia |

| Short form | Long form | |
|---|---|---|
| *Corrected pairs – continued* | | |
| SSSS | staphylococcal scalded skin syndrome | scalded skin syndrome |
| EBER | EBV-encoded small nuclear RNA | ed ebv-encoded small nuclear rna |
| HD | Hodgkin's disease | 15 with Hodgkin's disease (HD † |
| GluR | glutamate receptor | g chemical selectivity of agonists for the nmda subtype of glutamate receptor |
| TUNEL | terminal deoxynucleotidyl transferase mediated deoxyuridine triphosphate biotin nick end labelling | ted deoxyuridine triphosphate biotin nick end labelling |
| LC/ESI/MS/MS | HPLC/electrospray ionization tandem mass spectrometric | lective hplc/electrospray ionization tandem mass spectrometric |
| CYSP | cysteine peptide | conformations of the polypeptides beta endorphin |
| ESI/MS | electrospray ionization mass spectrometry | ectrospray ionization mass spectrometry |
| Lid | Lidocaine | lidated for the quantitation of lidocaine |
| DEX-MPS | dextran-methylprednisolone succinate | DEX-MPS) and its degradation products methylpr † |
| UV | Ultraviolet | 60:40 v/v) and ultraviolet (UV) detection at † |

† Incorrect short form

## 3 RESULTS

In the BioText corpus, we found 13 incorrectly matching or malformed 'id' attributes and 21 mismatched or malformed `<Long>` and `<Start>` tags. The corrected labelled corpus is available from http://soi.city.ac.uk/~abdy181/software/#badrex (reproduced with permission).

Table 1 shows the results of analysis of the Medstract corpus; against the Medstract gold standard markables: we added an additional 43 markables that we judged to be correct short-form–long-form pairs and amended 17 pairs that we judged to have incorrect spans. The corrected markables file is also available from the above URL.

Evaluation of BADREX performance against both corpora in comparison to that of S-H, ALICE and MBA are shown in Table 2.

**Table 2.** Evaluation results against corrected gold standard data sets

| System | Corpus | Precision | Recall | $F_1$ |
|---|---|---|---|---|
| BADREX[a] | Medstract | 98% | 97% | 0.97 |
| | BioText† | 90% | 85% | 0.88 |
| S-H | Medstract | 90% | 97% | 0.93 |
| | BioText | 91% | 79% | 0.85 |
| ALICE | Medstract | 98% | 94% | 0.96 |
| | BioText | 92% | 68% | 0.78 |
| MBA* | Medstract | 91% | 88% | 0.89 |
| | BioText | - | - | - |

a Coreference mode disabled in BADREX for this evaluation
† Corrected BioText corpus available in XML format from the author.
* Results reported by authors; software unavailable to evaluate on BioText corpus.

## 4 DISCUSSION

Our goal was to develop a customisable tool for identifying, expanding and annotating *in situ* biomedical abbreviations in free text, while matching or exceeding the performance of existing approaches. Running both 'head' and 'tail' candidate matches, allowing a variable threshold and only considering alpha characters when matching allows us to identify pairs such as 'topoisomerase I (Top1p)' and 'two-dimensional polyacrylamide gel electrophoresis (2D-PAGE)' missed by the other approaches, yet without unduly compromising precision.

By using regular expressions in steps 1 and 3, we simplify the creation of the finite state machine hard-coded in Schwartz & Hearst (2003) and allow it to be easily parameterised, so that optimal parameter values for a given corpus can be identified. While we used default values for these (detailed above) in this evaluation, iterative regression techniques could be used to find minimum values for the `maxOuterWords`, `maxInnerChars`, and `threshold` parameters that maximise precision and/or recall for other corpora.

We have not evaluated the coreferencing features of BADREX here. In addition, BADREX can be configured to annotate common medical abbreviations extracted from Wikipedia (Wikipedia 2012). Future work will evaluate the contribution of these features as components in the disambiguation of unpaired abbreviations. BADREX is currently coupled to GATE but a standalone version is planned.

## 5 CONCLUSION

The use of regular expressions dynamically generated from document content yields modestly improved performance over previous approaches to identifying term definition–term abbreviation pairs, with the benefit of providing in-place annotation, expansion and coreference in a single pass. BADREX requires no training data and allows runtime customisation of its input parameters. The GATE plugin is freely available from https://github.com/philgooch/BADREX-Biomedical-Abbreviation-Expander and is licensed under the GPLv3.

## ACKNOWLEDGEMENTS

We thank Prof. Marti Hearst for permission to make available the corrected BioText yeast corpus in XML format. The author acknowledges funding and support from the Engineering and Physical Sciences Research Council (EPSRC) in carrying out this research as part of PhD studentship EP/P504872/1.